\documentclass{article}

\usepackage[preprint]{neurips_2026}

\usepackage[utf8]{inputenc} %
\usepackage[T1]{fontenc}    %
\usepackage{hyperref}       %
\usepackage{url}            %
\usepackage{booktabs}       %
\usepackage{amsfonts}       %
\usepackage{nicefrac}       %
\usepackage{microtype}      %
\usepackage{xcolor}         %

\usepackage{lineno}

\definecolor{darkblue}{rgb}{0, 0, 0.5}
\hypersetup{colorlinks=true, citecolor=darkblue, linkcolor=darkblue, urlcolor=darkblue}

\usepackage{graphicx,xcolor}  %
\usepackage[whole]{bxcjkjatype}

\usepackage{wrapfig}

\hypersetup{
	colorlinks=true, 
    citecolor=blue, 
    linkcolor=blue,
    urlcolor=blue,
	pdfborder={0 0 0},
}

\usepackage{mathtools}

\usepackage{multirow}
\usepackage{cite}
\usepackage{breakcites}
\usepackage{amsfonts}

\usepackage{listings}
\usepackage{tcolorbox}
\tcbuselibrary{skins, breakable, listings}
\usepackage{tikz}
\usepackage{comment} %

\newtcolorbox[auto counter, number within=section]{Prompt}[2][]{%
  colback=white, %
  width=\linewidth, %
  arc=3mm, 
  boxrule=0.8mm, %
  title=\large #2, %
  breakable, %
  fonttitle=\small, %
  fontupper=\footnotesize, %
  #1 %
}

\newtcolorbox{SlimCode}{
  enhanced,
  colback=gray!20,
  colframe=gray!50,
  boxrule=0.4pt,
  arc=2mm,
  outer arc=2mm,
  left=2mm,
  right=2mm,
  top=1mm,
  bottom=1mm,
  boxsep=2pt,
  breakable,
  listing options={
    basicstyle=\ttfamily\scriptsize,  %
    breaklines=true,
    columns=fullflexible,
  },
  before skip=5pt,   %
  after skip=5pt     %
}

\AtBeginEnvironment{SlimCode}{\footnotesize}

\newcommand{\method}{JAMMEval}

\title{JAMMEval: A Refined Collection of Japanese Benchmarks for Reliable VLM Evaluation}

\author{Issa Sugiura$^{1,2}$
  Koki Maeda$^{5,2}$
  Shuhei Kurita$^{3,2}$
   Yusuke Oda$^{2}$\\
   {\bf Daisuke Kawahara$^{4,2}$  
  Naoaki Okazaki$^{5,2}$} \\
  $^1$Kyoto University 
  $^2$NII LLMC 
  $^3$NII 
  $^4$Waseda University
  $^5$Institute of Science Tokyo
}

\begin{document}

\maketitle

\begin{abstract}
Reliable evaluation is essential for the development of vision-language models (VLMs). However, Japanese VQA benchmarks have undergone far less iterative refinement than their English counterparts. As a result, many existing benchmarks contain issues such as ambiguous questions, incorrect answers, and instances that can be solved without visual grounding, undermining evaluation reliability and leading to misleading conclusions in model comparisons.
To address these limitations, we introduce \textbf{JAMMEval}, a refined collection of Japanese benchmarks for reliable VLM evaluation. It is constructed by systematically refining seven existing Japanese benchmark datasets through two rounds of human annotation, improving both data quality and evaluation reliability.
In our experiments, we evaluate open-weight and proprietary VLMs on \method{} and analyze the capabilities of recent models on Japanese VQA. We further demonstrate the effectiveness of our refinement by showing that the resulting benchmarks yield evaluation scores that better reflect model capability, exhibit lower run-to-run variance, and improve the ability to distinguish between models of different capability levels.
We release our dataset and code to advance reliable evaluation of VLMs.\footnote{\url{https://speed1313.github.io/JAMMEval}}
\end{abstract}

\section{Introduction}

Vision–language models (VLMs), which process both images and text, have made rapid progress~\citep{liu2023llava,wang2025internvl35,bai2025qwen3vl,openai2025gpt5.1,google2025gemini3pro}. Reliable evaluation plays a central role in model development, informing key decisions such as the choice of training dataset and model architectures, and ultimately shaping the direction of progress~\citep{penedo2025fineweb2,jason2025eval}.

However, constructing reliable evaluation datasets is challenging in practice, and noise can be introduced during dataset construction, potentially distorting evaluation results~\citep{chen2024mmstar,joshi2026datbench}. Such noise can arise in several forms. For example, ambiguous questions or incorrect reference annotations can cause correct model outputs to be judged as incorrect, artificially lowering evaluation scores. In addition, some questions can be answered using only the textual question without consulting the image, producing evaluation signals that do not reflect the model's visual understanding~\citep{chen2024mmstar}.

\begin{figure}[t]
\begin{center}
\includegraphics[width=\linewidth]{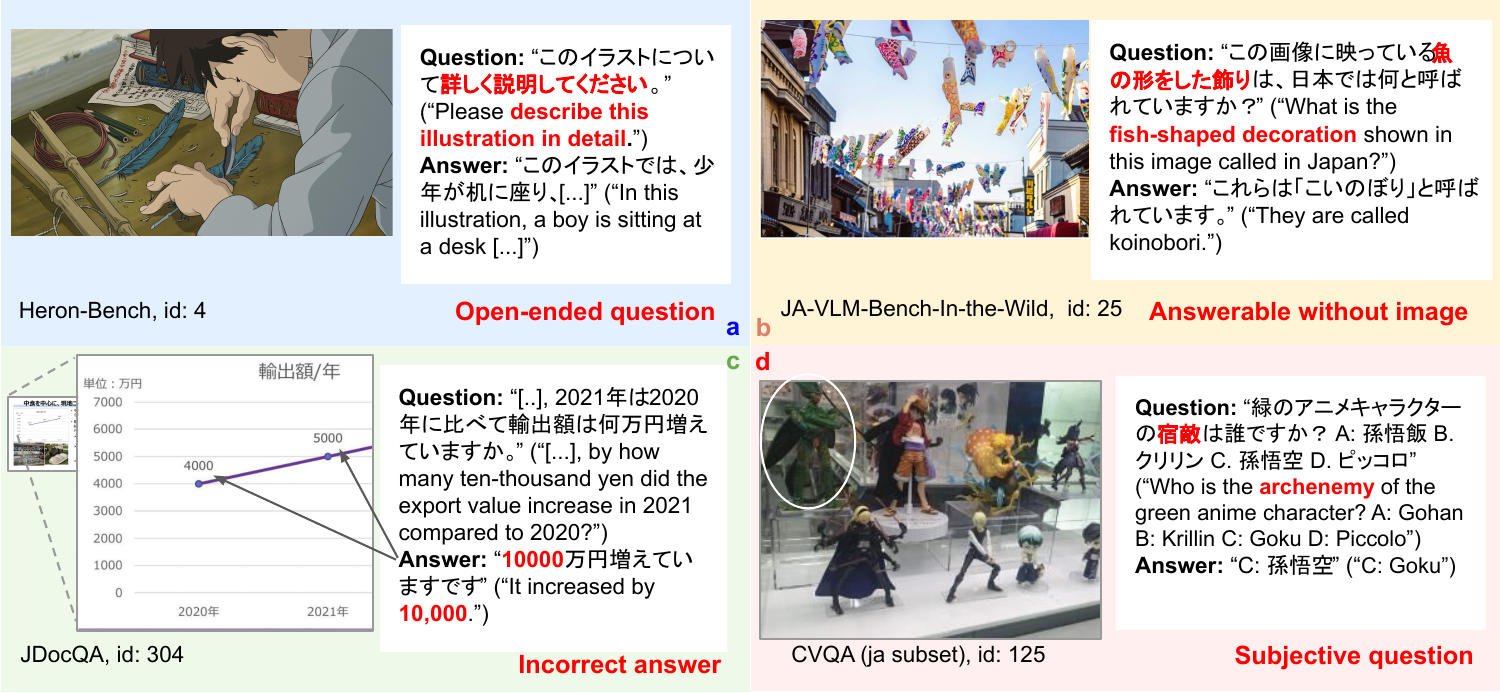}
\caption{Examples of inappropriate instances in existing Japanese VQA evaluation datasets. (a) Open-ended questions with inherent ambiguity that do not admit a unique correct answer. (b) Questions that can be answered with high confidence using only the text, without referring to the image. (c) Instances with incorrect ground-truth answers. (d) Subjective questions for which answers vary across annotators.}
\label{fig:errors}
\end{center}
\end{figure}

Many English VQA benchmarks have undergone iterative refinement across diverse domains~\citep{chen2024mmstar,yue-etal-2025-mmmu,joshi2026datbench,zhang2024mathverse,wang2024charxiv}.
In contrast, Japanese VQA benchmarks have undergone far less iterative refinement.As a result, our manual inspection reveals that many datasets still contain issues, such as those illustrated in Figure~\ref{fig:errors}, including ambiguous questions, incorrect answers, and instances that can be solved without visual grounding in the image. These issues undermine the reliability of VLM evaluation.

To address these issues, we introduce \textbf{\method{}} (\textbf{JA}panese \textbf{M}ulti\textbf{M}odal \textbf{Eval}uation Collection), a curated benchmark collection for evaluating VLMs on Japanese VQA.
JAMMEval is constructed by refining seven existing Japanese VQA benchmarks across diverse domains, such as cultural knowledge, charts, documents, and OCR-based question answering, through two rounds of manual review and re-annotation.
In our experiments, we evaluate recent open-weight and proprietary VLMs on JAMMEval to assess the current state of Japanese multimodal understanding in VLMs.
We further show that our refinement process produces evaluation scores that more faithfully reflect model capability, exhibit lower run-to-run variance, and improve model distinguishability, highlighting the importance of benchmark quality in obtaining reliable and informative evaluations.

We will release our dataset and code to support future research on reliable VLM evaluation.

\section{Related Works}

\noindent\textbf{Benchmark datasets for vision-language models.}
Benchmark datasets have played a central role in the development and evaluation of vision–language models (VLMs).
Early VLM research primarily evaluated models on a limited set of tasks such as natural image captioning or basic visual question answering~\citep{liu2023llava}.
As VLMs have grown more capable, recent benchmarks have expanded to cover a broader range of capabilities, recent benchmarks have expanded to cover a broader range of capabilities, including chart and table understanding, document reasoning, OCR-based tasks, and agentic tasks such as computer use~\citep{yue2023mmmu,masry2022chartqa,wang2024charxiv,zhang2024mathverse,li2025screenspotpro,xie2024osworld}.
Several Japanese VQA benchmarks have also been proposed to evaluate the Japanese multimodal performance of VLMs, covering diverse domains such as Japanese cultural knowledge, documents, charts, and slides~\citep{inoue2024heronbench,onami2024jdocqa,jgraphqa,stockmark2025businessslidevqa}.

\noindent\textbf{Benchmark refinement.}
Constructing reliable evaluation datasets remains challenging~\citep{anthropic2026demystifyingevals,jason2025eval}.
To improve reliability, several approaches have been proposed to refine and curate existing benchmarks~\citep{chen2024mmstar,yue-etal-2025-mmmu,wang2024charxiv,zhang2024mathverse}.
MMStar~\citep{chen2024mmstar} improves dataset quality by filtering 22,401 instances from six English benchmarks down to 1,500 high-quality examples via LLM-based coarse filtering followed by manual review.
DatBench improves benchmark quality by converting multiple-choice questions into a generative format that is less susceptible to guessing. It also filters out problematic instances, including those solvable without images, ambiguous questions, incorrect answers, and cases where images are too low-resolution to support reliable reasoning.

Despite these advances, most benchmark refinement efforts have focused on English datasets, while Japanese VQA benchmarks remain relatively under-refined, posing a barrier to the development of Japanese VLMs~\citep{sasagawa-etal-2025-constructing,maeda2025llm-jp-eval-mm,sugiura2025waon}.
This imbalance is partly due to the smaller research community surrounding Japanese benchmarks compared to English.
Moreover, prior work on English benchmarks typically assumes a sufficiently large sample size, allowing quality improvements primarily through filtering noisy instances~\citep{chen2024mmstar,joshi2026datbench}. In contrast, many Japanese benchmarks contain only a small number of samples (sometimes fewer than 100), making aggressive filtering problematic as it reduces dataset size and increases evaluation variance.
To address this limitation, our approach emphasizes re-annotation of noisy instances, enabling us to preserve the sample size while improving overall data quality.

\begin{figure}[t]
\begin{center}
\includegraphics[width=\linewidth]{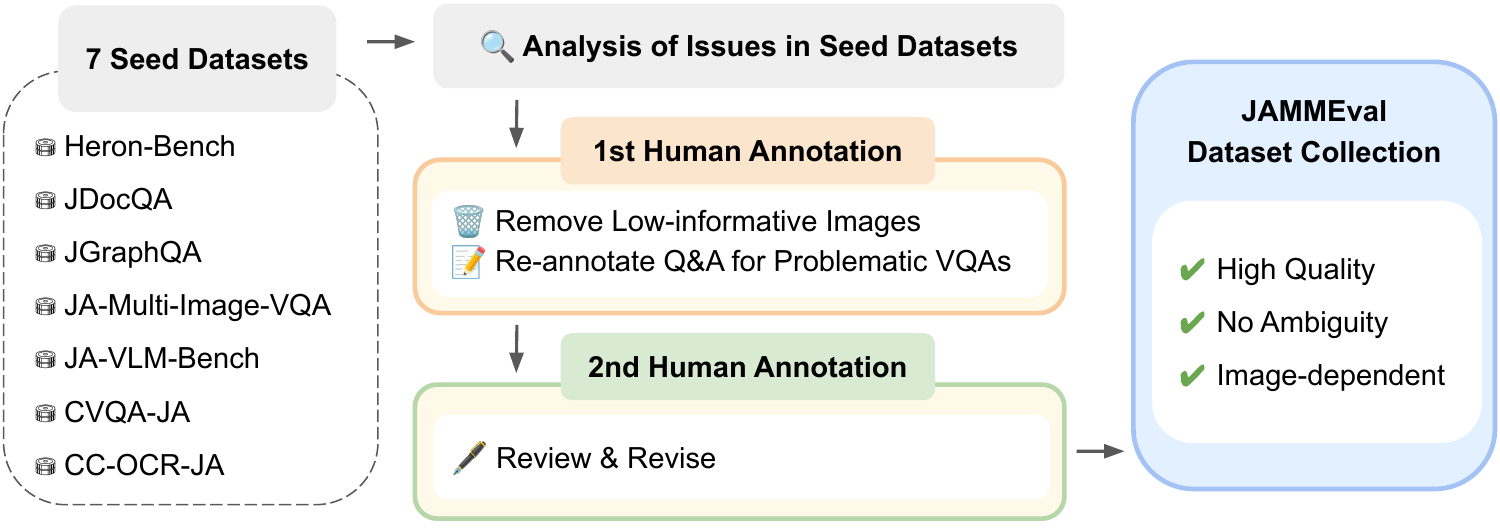}
\caption{Construction pipeline of \method{}. Starting from seven seed datasets, all instances undergo two rounds of manual review and re-annotation to produce a refined benchmark collection.}
\label{fig:JAMMEval-pipeline}
\end{center}
\end{figure}

\section{Construction of \method{}}
\label{sec:method}

Figure~\ref{fig:JAMMEval-pipeline} illustrates the construction pipeline of \method{}. We construct \method{} by first selecting seven seed Japanese VQA datasets, analyzing common issues in existing benchmarks through an initial review of all instances, and then refining them through a two-round manual review and re-annotation process.

\subsection{Seed Dataset Selection}

We select seven seed datasets from existing Japanese VQA benchmarks spanning diverse domains, including OCR, Japanese cultural knowledge, multi-image, document, and chart \& table. In total, the seed datasets contain 1,925 instances.

\noindent\textbf{CC-OCR-JA}~\citep{yang2024ccocr} is the Japanese subset of CC-OCR. The dataset consists of natural images taken in Japan, such as signboards and road signs containing Japanese text. The task requires extracting all readable text from an image in reading order.

\noindent\textbf{JA-VLM-Bench-In-the-Wild}~\citep{akiba2025evo} is a VQA dataset composed of images depicting Japanese culture and objects commonly found in Japan.

\noindent\textbf{Heron-Bench}~\citep{inoue2024heronbench} is a VQA dataset comprising images related to Japanese culture and scenery, including scenes from Studio Ghibli films, Japanese architecture, and traditional paintings.

\noindent\textbf{CVQA-JA}~\citep{mogrovejo2024cvqa} is the Japanese subset of CVQA, a culturally diverse multilingual VQA benchmark with multiple-choice questions. The dataset includes questions requiring knowledge of Japanese cultural common sense and pop culture, such as anime.

\noindent\textbf{JA-Multi-Image-VQA}~\citep{inoue2024jamultiimage} is a dataset designed to evaluate the ability of VLMs to perform question answering over multiple images. Each instance requires reasoning across multiple images simultaneously, such as identifying which image contains a cat, making it unsolvable without the ability to jointly process multiple images.

\noindent\textbf{JDocQA}~\citep{onami2024jdocqa} is a VQA dataset constructed from document images released by Japanese public institutions. Since the images are derived from PDFs, they tend to be high resolution, often requiring models to read small text within the image.

\noindent\textbf{JGraphQA}~\citep{jgraphqa} is a Japanese chart and table understanding benchmark based on images extracted from Japanese investor relations (IR) materials. The dataset comprises four types of figures: pie charts, line charts, bar charts, and tables.

\subsection{Common Issues Found in Existing Japanese VQA Benchmarks}

Before performing dataset refinement, we first analyzed the instances in the selected seed datasets to identify common issues in existing benchmarks. Prior work on refining English VQA benchmarks has reported several types of inappropriate evaluation instances~\citep{chen2024mmstar,yue-etal-2025-mmmu,joshi2026datbench}. Through manual inspection of the seed datasets, we found that similar issues also appear in the Japanese benchmarks. In particular, we identified the following typical issues, with representative examples shown in Figure~\ref{fig:errors}:
\begin{itemize}

  \item  \noindent\textbf{Ambiguity.} Questions that do not admit a single unambiguous correct answer. This includes open-ended questions that invite multiple valid responses and subjective questions for which answers may vary across respondents. Without a single definitive answer, correctness judgments depend on the evaluator's interpretation, introducing subjectivity into grading.

  \item \noindent\textbf{Lack of image grounding.} Questions that can be answered without referring to the image. Since VQA benchmarks are intended to measure a model's visual understanding, instances that can be resolved from textual cues alone do not serve this purpose.

  \item \noindent\textbf{Incorrect ground-truth answers.} Instances where the annotated answer is incorrect. When a model produces the genuinely correct answer, it is penalized rather than rewarded, directly suppressing evaluation scores below their true value.
\end{itemize}

\subsection{Dataset Refinement}
We refined the seed datasets through two rounds of human annotation.

\noindent\textbf{Answer Format.}
We first define a unified answer format for each dataset. For CVQA-JA, we retain the original multiple-choice format. For all other datasets, we standardize the format to short-answer questions, a format widely adopted in English VQA benchmarks~\citep{masry2022chartqa,Mathew2021docvqa,Mathew2022infovqa}, where models produce short responses such as a single word or phrase. Most seed datasets already follow a short-answer style. A notable exception is CC-OCR-JA, where answers were originally structured as reading-order extractions. We convert these into short-answer format to enable consistent and comparable evaluation across all datasets under a unified protocol.

\noindent\textbf{Annotation Process.}
In each round, annotators reviewed existing VQA instances for issues such as ambiguity, incorrect ground-truth answers, or insufficient visual grounding, and re-annotated problematic instances accordingly. While prior work such as MMStar~\citep{chen2024mmstar} handled problematic cases by removing them, we prioritized revision over deletion, since some seed datasets comprised fewer than 100 examples in total. We clarified ambiguous QA pairs, corrected erroneous answers, and when minor edits were insufficient, created new QA pairs from the same image. Some datasets also required additional modifications to address dataset-specific issues, which we describe in Appendix~\ref{sec:dataset-specific-modifications}.
The first round of manual annotation was conducted by the authors, who reviewed and refined the seed datasets. The second round employed external annotators to catch issues missed in the first round and to fix minor problems such as typos or incorrect modifications introduced during the first round.

\begin{figure}[t]
\begin{center}
\includegraphics[width=\linewidth]{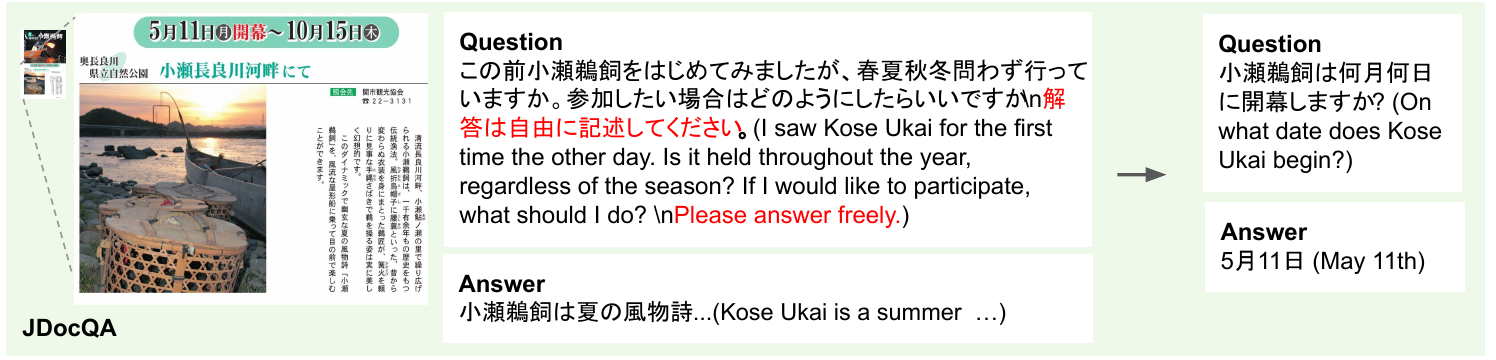}
\caption{An example of re-annotation. An ambiguous open-ended question is replaced with a specific, objectively answerable question targeting information visible in the image.}
\label{fig:annotation-example}
\end{center}
\end{figure}
Figure~\ref{fig:annotation-example} illustrates an example of re-annotation. The original question was ambiguous due to its open-ended formulation (``Please answer freely''). The refined version asks for the specific opening date shown in the image, removing ambiguity.

\begin{figure}[t]
\begin{center}
\includegraphics[width=0.9\linewidth]{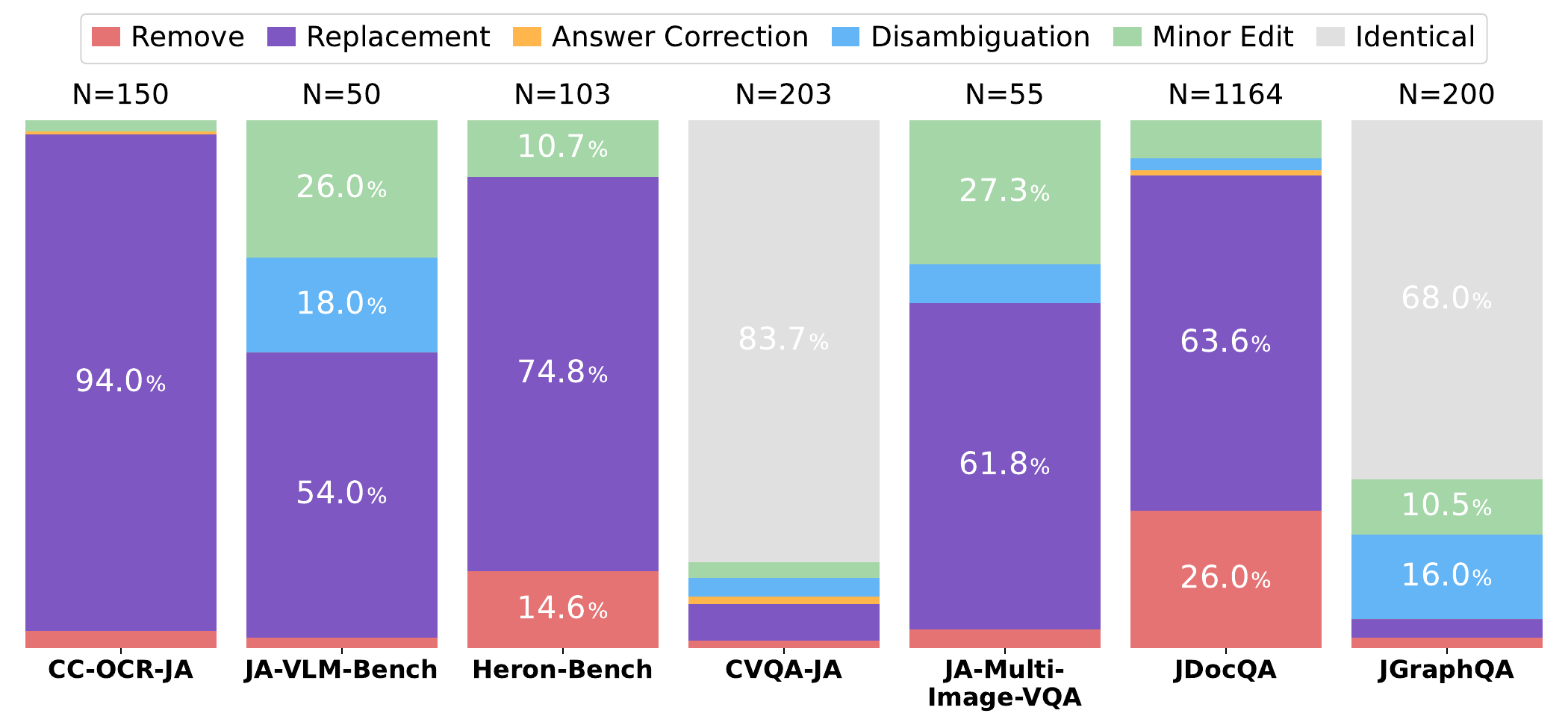}
\caption{Breakdown of refinement operations per dataset. Identical instances required no modification; other categories indicate the type of correction applied.}
\label{fig:reannotation-stats}
\end{center}
\end{figure}

\subsection{Distribution of Refinement Operations Across Datasets}

To analyze how refinement operations were applied to individual instances, we categorize each instance into one of six types:
\begin{enumerate}
    \item \textbf{Identical}: instance retained as-is with no issues identified.
    \item \textbf{Minor edit}: minor wording, formatting, or typographical fixes with no semantic impact.
    \item \textbf{Disambiguation}: question or answer clarified to remove ambiguity while preserving the original intent.
    \item \textbf{Answer correction}: answer corrected while the question remains essentially unchanged.
    \item \textbf{Replacement}: a new question–answer pair created because the original instance contains issues that cannot be resolved through minor edits or clarification.
    \item \textbf{Remove}: instance removed due to insufficient visual information or redundancy.

\end{enumerate}

Figure~\ref{fig:reannotation-stats} shows the distribution of refinement operations across the seven seed datasets in \method{}. 
Compared to other datasets, JGraphQA and CVQA-JA have a notably higher proportion of instances in the Identical category, with more than half retained without modification. This can be attributed to the fact that both datasets incorporate a verification stage during construction, which ensures a certain level of quality prior to release. This highlights the importance of validation during dataset construction in ensuring data quality.

Heron-Bench and JDocQA contain a relatively large fraction of removed instances (14.6\% and 26.0\%, respectively). In the original Heron-Bench dataset, five QA pairs are provided per image, which made it challenging for annotators to consistently construct unambiguous and meaningful questions for every image, leading to frequent removals. JDocQA contains a number of visually similar or visually uninformative images (e.g., title pages of reports), which posed a similar challenge for annotators.

It is worth noting that the high proportion of Replacement instances in CC-OCR-JA is not indicative of quality issues in the original dataset, but rather reflects the format conversion from its original text extraction style to the short-answer format.

\begin{table}[t]
\centering
\small
\setlength{\tabcolsep}{3pt}
\caption{Statistics of \method{}. The collection spans seven tasks across diverse domains, totaling 1,592 instances after refinement.}
\label{tab:dataset_stats}
\begin{tabular}{llrrccc}
\toprule
\textbf{Dataset} &  \textbf{Seed Dataset} & \textbf{Samples} & \textbf{Images} & \textbf{Answer Format} & \textbf{Category}\\
\midrule
CC-OCR-JA-Refined &{\scriptsize \citet{yang2024ccocr}} & 145 &145  &Short answer & OCR\\
JA-VLM-Bench-Refined  &{\scriptsize \citet{akiba2025evo}}& 49 & 42  & Short answer & Japanese culture\\
Heron-Bench-Refined& {\scriptsize \citet{inoue2024heronbench}} &	88 & 21  & Short answer & Japanese culture \\
CVQA-JA-Refined &{\scriptsize \citet{mogrovejo2024cvqa}}& 200  & 94 & Multiple-choice & Japanese culture\\
JA-Multi-Image-VQA-Refined  &{\scriptsize \citet{inoue2024jamultiimage}}& 53&  39 & Short answer& Multi-image\\
JDocQA-Refined &{\scriptsize \citet{onami2024jdocqa}}& 861 & 793 &Short answer & Document\\
JGraphQA-Refined &{\scriptsize \citet{jgraphqa}} & 196 & 98 & Short answer & Chart \& Table\\
\midrule
Total & -- & 1,592 & 1,232 & -- & -- \\
\bottomrule
\end{tabular}
\end{table}

\subsection{Statistics of \method{}}

Table~\ref{tab:dataset_stats} presents the statistics of \method{}. To distinguish the refined datasets from their originals, we append ``-Refined'' to each dataset name. As a result of the refinement process, \method{} comprises 1,592 VQA instances in total, reduced from 1,925 instances in the seed datasets.

\section{Model Evaluation on \method{}}
\label{sec:eval}
In this section, we evaluate existing models on \method{} to benchmark the current state of Japanese multimodal understanding.

\subsection{Evaluation Setup}
\noindent\textbf{Models.}
We evaluate both open-weight and proprietary models. For open-weight models, we include the multilingual Qwen3-VL-\{2B, 4B, 8B\}~\citep{bai2025qwen3vl} and InternVL3.5-\{2B, 4B, 8B\}~\citep{wang2025internvl35}, and Sarashina2.2-Vision-3B~\citep{sbintuitions2025sarashina}, a Japanese-specialized model. For proprietary models, we evaluate GPT-4o (\texttt{gpt-4o-2024-11-20})~\citep{openai2024gpt4ocard}, GPT-5.1 (\texttt{gpt-5.1-2025-11-13})~\citep{openai2025gpt5.1}, and Gemini~3~Pro (\texttt{gemini-3-pro-preview})~\citep{google2025gemini3pro}.

\noindent\textbf{Generation settings.}
To encourage models to respond in the defined answer format, we use format-specific prompts for short-answer and multiple-choice questions, respectively. The full prompt templates are provided in Appendix~\ref{sec:prompt_template}.
We set the temperature to 0 for all models.
The maximum number of generated tokens is set to a value sufficient for each dataset.
For GPT-4o and GPT-5.1, the detail parameter was set to high.
For GPT-5.1, we set the reasoning effort to \texttt{none}.
For Gemini~3~Pro, we set the reasoning level to \texttt{low} and the maximum number of tokens to 1,024.
All open-weight models were run on NVIDIA H200 GPUs.

\noindent\textbf{Evaluation Protocol.}
We use accuracy as the evaluation metric across all datasets. For short-answer questions, we adopt LLM-based soft exact match to account for minor variations in model outputs, such as differences in full-width and half-width characters or units. Specifically, the judge model is provided with a triplet consisting of the question, the reference answer, and the model's output, and is tasked with making a binary judgment (correct or incorrect); the grading prompt is provided in Appendix~\ref{sec:grading_prompt}. We use GPT-5.1 as the judge model.
For multiple-choice questions, the predicted option letter is extracted via regular expression and evaluated by exact match. To account for potential non-determinism in both the evaluated and judge models, we conducted three evaluation runs and reported the mean and standard deviation.

\begin{figure}[t]
\begin{center}
\includegraphics[width=\linewidth]{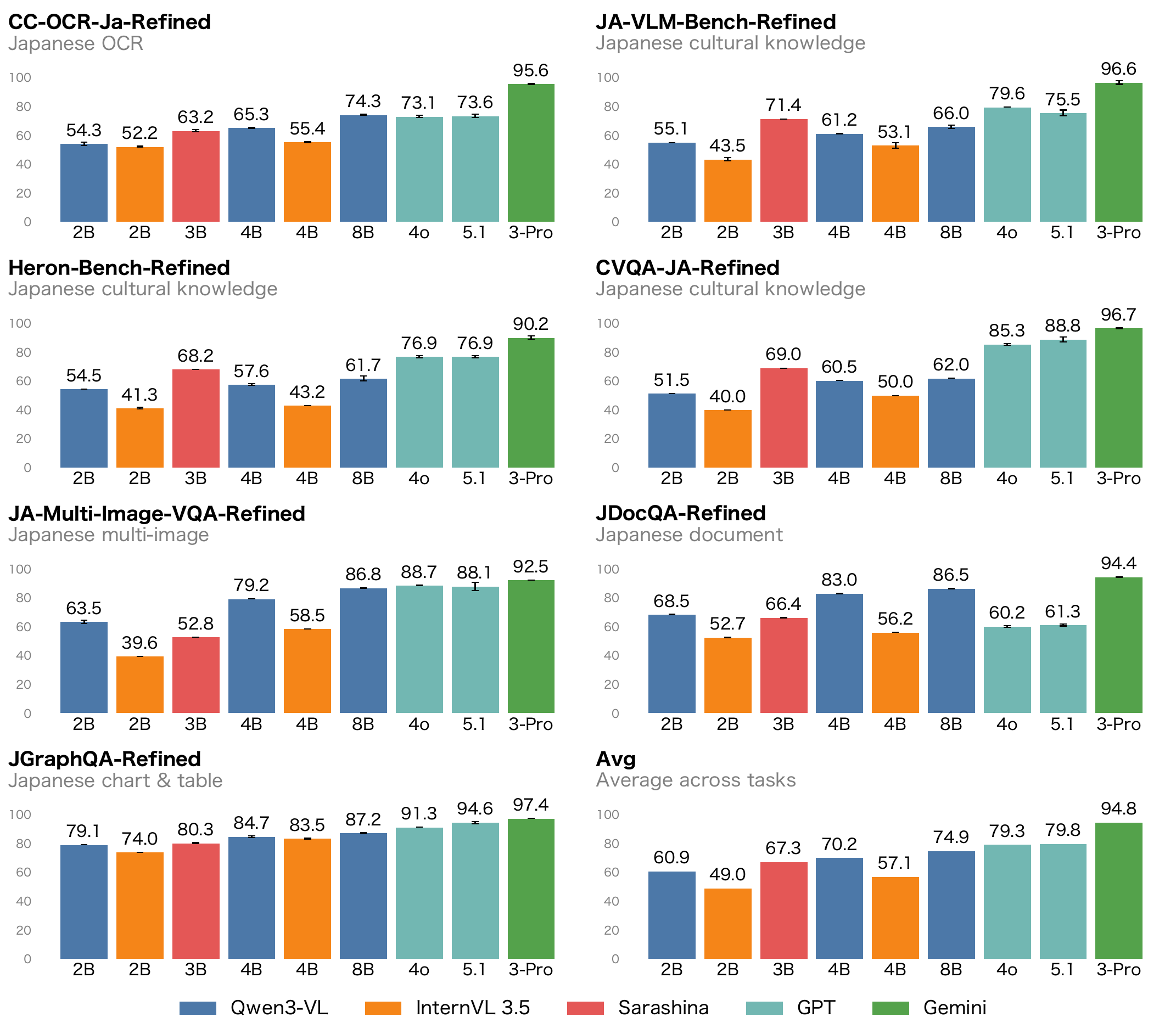}
\caption{Model performance on JAMMEval across seven tasks. Note that Gemini 3 Pro is evaluated with reasoning enabled, while all other models are evaluated without reasoning. Gemini 3 Pro achieves the highest scores overall.}
\label{fig:result}
\end{center}
\end{figure}

\subsection{Evaluation Results}
The evaluation results are shown in Figure~\ref{fig:result}.

\noindent\textbf{Gemini 3 Pro leads overall.} It exceeds 90\% accuracy on every task in \method{}, demonstrating strong generalization across diverse domains. Note that Gemini~3~Pro is evaluated with reasoning enabled, while all other models are evaluated without reasoning; enabling reasoning for other models such as GPT-5.1 and Qwen3-VL might yield comparable performance.

\noindent\textbf{Qwen3-VL-8B leads among open-weight models.} It achieves strong performance across a wide range of tasks, and notably outperforms GPT-5.1 on text-reading tasks such as JDocQA-Refined and CC-OCR-JA-Refined.

\noindent\textbf{Sarashina2.2-Vision-3B excels at Japanese cultural tasks.} It outperforms similarly sized models on Japanese cultural tasks, including Heron-Bench-Refined, JA-VLM-Bench-Refined, and CVQA-JA-Refined, suggesting the effectiveness of Japanese-specific pretraining.

\subsection{Error Analysis of Gemini 3 Pro}
\begin{figure}[t]
\begin{center}
\includegraphics[width=0.9\linewidth]{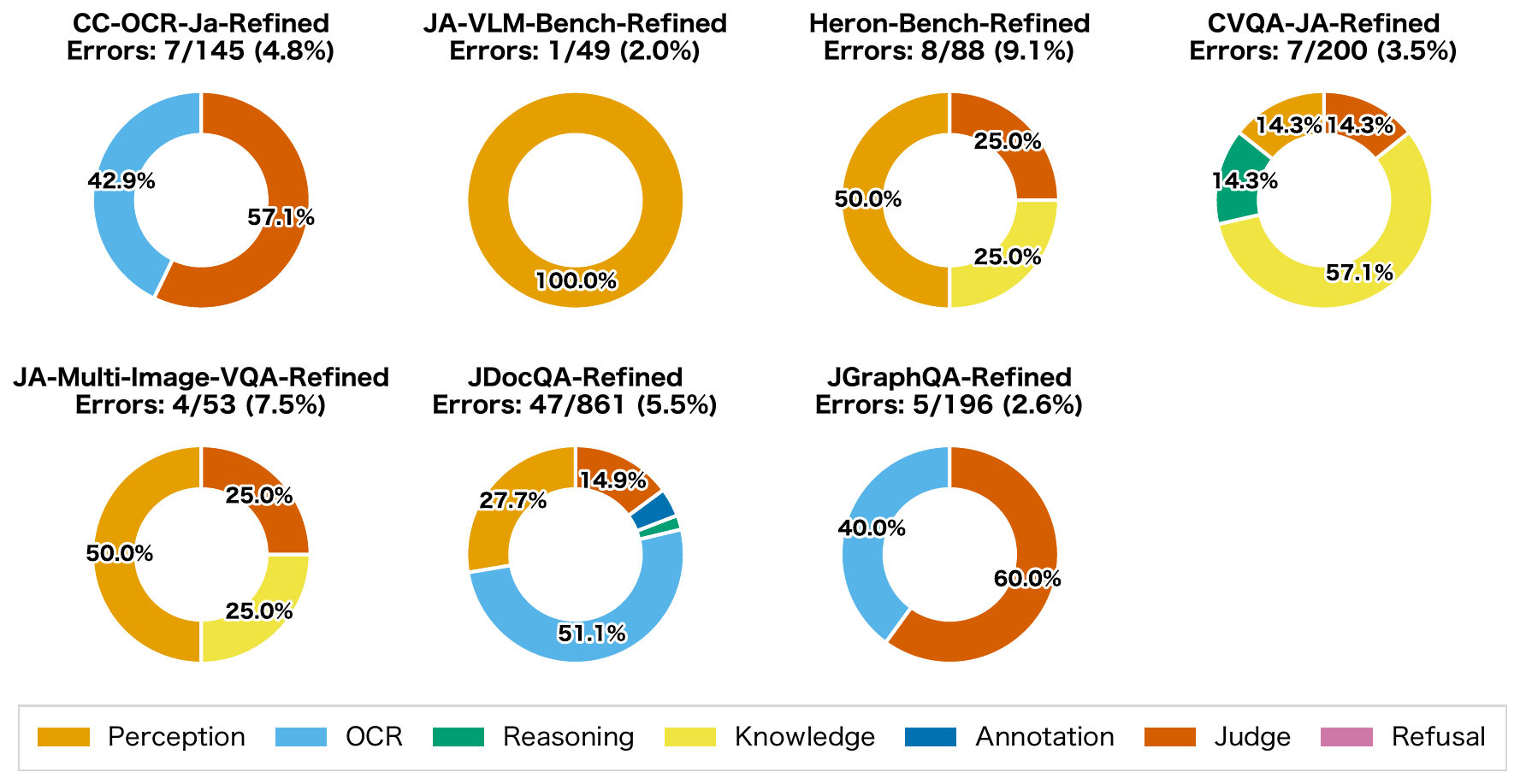}
\caption{Distribution of Gemini~3~Pro errors by category across datasets. Knowledge errors are prominent in CVQA-JA-Refined, reflecting limited Japanese cultural knowledge, while perception errors appear broadly across datasets. Judge errors in short-answer datasets reflect limitations of LLM-based grading rather than model capability.}
\label{fig:gemini3-error-stats}
\end{center}
\end{figure}

We analyze errors made by the best-performing model in our evaluation, Gemini~3~Pro, to identify challenges for further improving model performance. We manually categorized these errors into seven types, following the error taxonomy used in \citet{yue-etal-2025-mmmu} with modifications to suit our setting: Perception, OCR, Reasoning, Knowledge, Judge, Annotation, and Refusal.
Figure~\ref{fig:gemini3-error-stats} summarizes the error distribution for Gemini~3~Pro, and representative error cases are provided in Appendix~\ref{app:error_cases}.

\noindent\textbf{Knowledge Errors.}
CVQA-JA-Refined, which evaluates Japanese cultural knowledge, exhibits a high proportion of knowledge errors, suggesting that Gemini~3~Pro still lacks sufficient knowledge of Japanese culture and has room for further improvement.

\noindent\textbf{Perception Errors.}
Perception errors are observed across a wide range of datasets, mainly caused by hallucinations in tasks such as object counting, left–right confusion (e.g., misidentifying the left and right foot), and misinterpreting spatial relationships within images.

\noindent\textbf{Judge Errors.}
Datasets with a short-answer format exhibit some judge errors due to limitations of LLM-based grading. Because the judge model evaluates responses without visual context, it may occasionally penalize semantically correct answers that deviate from the reference wording, despite their alignment with the image. These errors reflect limitations of the evaluation protocol rather than model capability. Recent work has explored VLM-as-Judge~\citep{dongping2024mllm-as-a-judge}, where the judge model is given access to the image during grading; incorporating such methods remains future work.

\begin{table}[t]
\centering
\small
\setlength{\tabcolsep}{3pt}
\caption{Evaluation statistics before and after refinement. \textbf{Accuracy} and \textbf{Run Std} denote the mean accuracy and the run-to-run standard deviation averaged across models per dataset. \textbf{Performance Gap} measures the score difference between the best and worst models. \textbf{Rank Corr.} denotes the Spearman rank correlation between model rankings before and after refinement. Refinement consistently raises accuracy and reduces run-to-run variance while preserving model rankings (Rank Corr.\ $\approx$ 1).}
\label{tab:mean_and_std_before_and_after}
\begin{tabular}{l ccc ccc cc c}
\toprule
 & \multicolumn{3}{c}{\textbf{Accuracy (\%)}} & \multicolumn{3}{c}{\textbf{Run Std}} & \multicolumn{2}{c}{\textbf{Performance Gap}} & \textbf{Rank Corr.}  \\
\cmidrule(lr){2-4} \cmidrule(lr){5-7} \cmidrule(lr){8-9}
\textbf{Dataset} & Before & After & $\Delta$ & Before & After & $\Delta$ & Before & After &\\
\midrule
CC-OCR-Ja & 22.4 & 65.8 & +43.4 & 0.9 & 0.6 & -0.3 & 26.0 & 44.6 & 0.78\\
JA-VLM-Bench & 64.1 & 64.9 & +0.8 & 1.6 & 0.8 & -0.9 & 28.7 & 53.1 & 0.88\\
Heron-Bench & 46.9 & 62.2 & +15.2 & 2.0 & 0.6 & -1.4 & 43.0 & 48.9 & 0.95\\
CVQA-JA & 62.6 & 65.3 & +2.7 & 0.9 & 0.8 & -0.1 & 58.0 & 56.7 & 0.96\\
JA-Multi-Image-VQA & 66.5 & 71.1 & +4.6 & 1.0 & 0.5 & -0.5 & 61.2 & 52.8 & 0.99\\
JDocQA & 42.5 & 68.8 & +26.3 & 0.3 & 0.2 & -0.1 & 35.3 & 41.7 & 0.77\\
JGraphQA & 82.5 & 85.7 & +3.2 & 0.6 & 0.3 & -0.3 & 23.2 & 23.5 & 1.00\\
\bottomrule
\end{tabular}
\end{table}

\begin{figure}[t]
\begin{center}
\includegraphics[width=\linewidth]{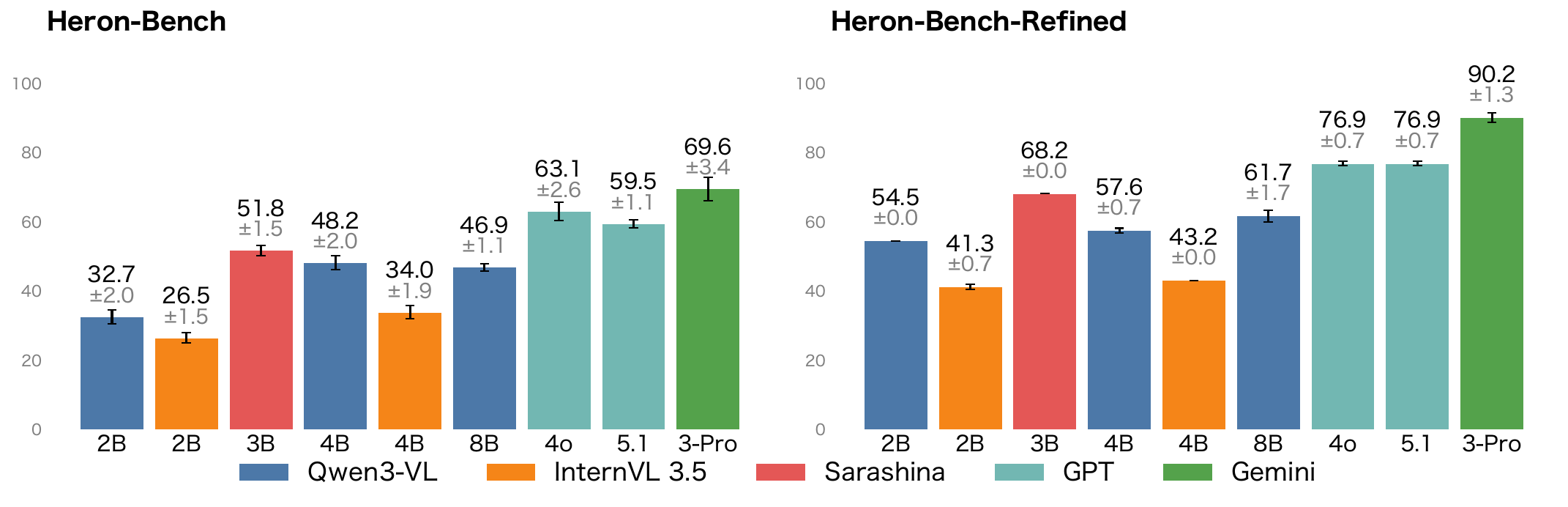}
\caption{Effect of refinement on Heron-Bench. After refinement, accuracy increases across all models and run-to-run variance decreases, likely because problematic instances are removed and ambiguous questions are replaced with objectively answerable ones, resulting in more stable and reliable evaluation scores.}
\label{fig:heronbench_comparison}
\end{center}
\end{figure}

\section{How Refinement Affects Benchmark Quality}
\label{sec:effect_of_dataset_refinement}
To examine how the refinement process affects benchmark quality, we evaluate models on both the original and refined versions of each dataset under the same evaluation setup described in Section~\ref{sec:eval}.

Table~\ref{tab:mean_and_std_before_and_after} summarizes evaluation statistics before and after refinement across datasets, reporting mean accuracy and run-to-run standard deviation averaged across models per dataset, along with the performance gap between the best and worst models and the Spearman rank correlation between model rankings before and after refinement.
As a representative example, Figure~\ref{fig:heronbench_comparison} compares benchmark results on Heron-Bench and Heron-Bench-Refined; results for the other datasets are provided in Appendix~\ref{sec:benchmark_results_with_old}.

\noindent\textbf{Model rankings are preserved.}
Model rankings remain largely consistent across the original and refined datasets, with the rank correlation close to 1 for most datasets, suggesting that the refinement process preserves the relative ordering of models.

\noindent\textbf{Accuracy increases.}
Accuracy scores are consistently higher on the refined versions, suggesting that problematic annotations in the original datasets, such as ambiguous questions and incorrect ground-truth answers, may have systematically underestimated model performance.

\noindent\textbf{Evaluation stability improves.}
The run-to-run standard deviation generally decreases after refinement, improving the signal-to-noise ratio of evaluation scores and making benchmark results more reliable and reproducible.

\begin{wrapfigure}{r}{0.48\textwidth}
\vspace{-4mm}
\centering
\includegraphics[width=\linewidth]{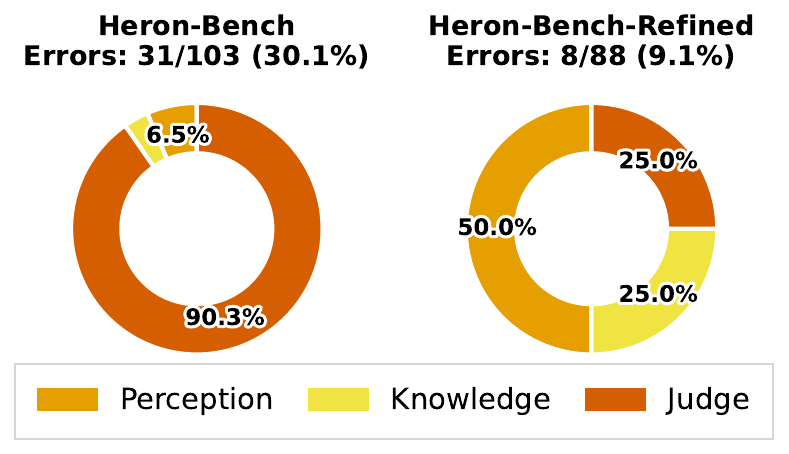}
\caption{Error distribution of Gemini 3 Pro on Heron-Bench and Heron-Bench-Refined.
In Heron-Bench, a large portion of errors are attributed to judge errors, primarily due to the prevalence of ambiguous QA pairs.} 
\label{fig:heronbench_error_stats}
\vspace{-4mm}
\end{wrapfigure}
\noindent\textbf{Model distinguishability improves.}
The performance gap between the best and worst models increases for most datasets (e.g., from 43.0 to 48.9 on Heron-Bench, and from 26.0 to 44.6 on CC-OCR-JA), suggesting that refinement improves the ability to distinguish between models of different capability levels. This increased sensitivity makes the refined datasets more suitable for fine-grained comparisons, such as ablation studies, where detecting subtle performance differences is critical.

To better understand the source of these improvements, Figure~\ref{fig:heronbench_error_stats} shows the distribution of error types for Gemini 3 Pro on Heron-Bench before and after refinement.
When the original dataset is used for evaluation, the overall error rate is higher, but the majority of errors are judge errors rather than errors attributable to the model itself.
These findings suggest that problematic instances in the original datasets impose an artificial ceiling on accuracy and impede objective grading, resulting in distorted and unstable evaluation signals.

\section{Conclusion}
We constructed JAMMEval, a refined Japanese VQA benchmark collection built by systematically refining seven existing benchmarks.
We evaluated a range of open and proprietary models on JAMMEval to assess the current level of multimodal performance on Japanese VQA tasks. 
We further demonstrated the effectiveness of our refinement process by showing that the refined benchmarks yield evaluation scores that better reflect model capability, exhibit lower run-to-run variance, and improve the ability to distinguish between models of different capability levels.

\section*{Acknowledgments}

We used ABCI 3.0 provided by AIST and AIST Solutions with support from ``ABCI 3.0 Development Acceleration Use''.

Furthermore, we would like to express our gratitude to the developers of the seed dataset used in the construction of \method{}.

\bibliography{neurips_2026}
\bibliographystyle{iclr2026_conference}

\appendix

\begin{figure}[t]
\begin{center}
\fbox{%
  \includegraphics[width=\linewidth]{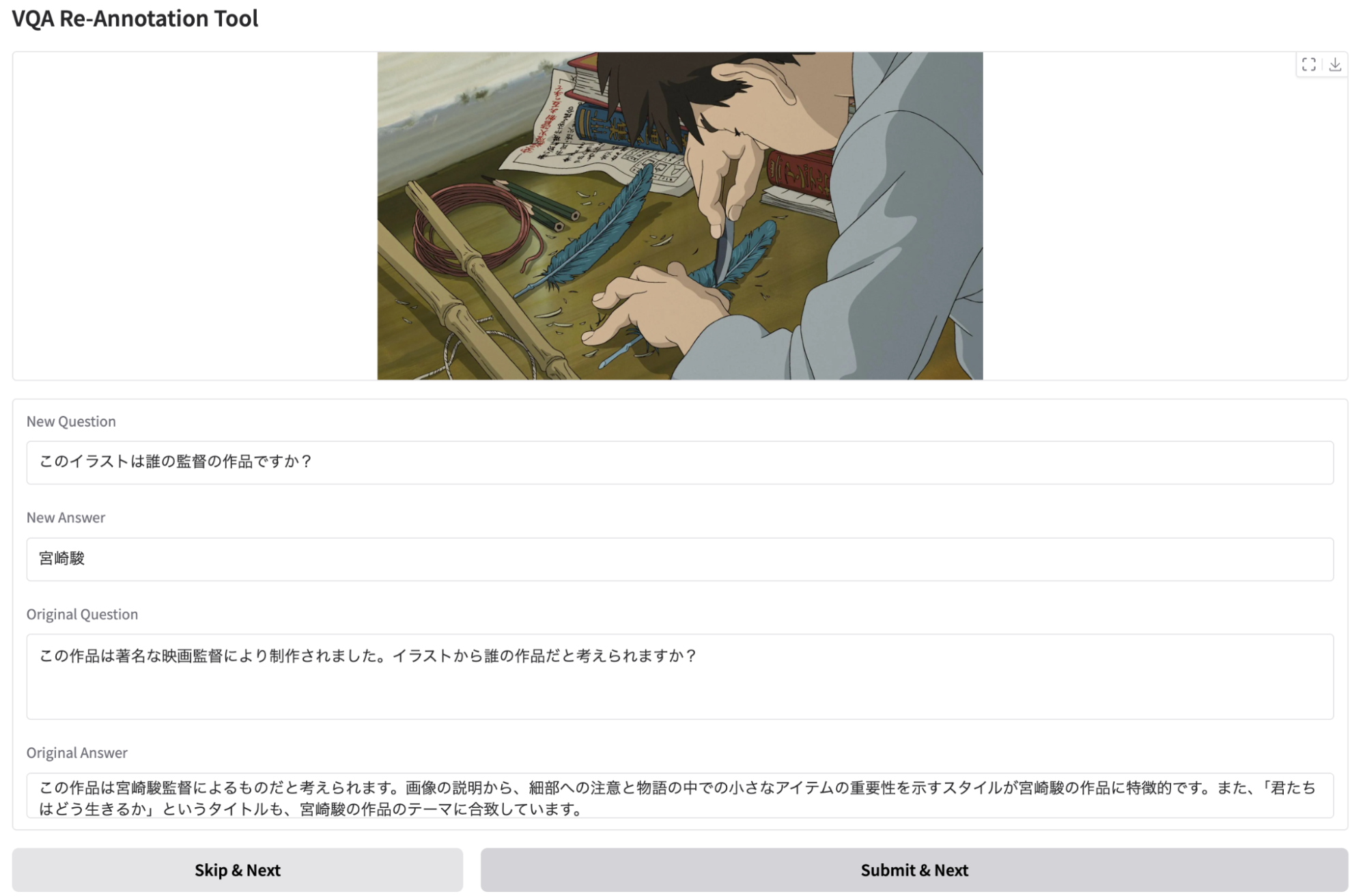}
}
\caption{Interface of the annotation tool. For each example, the (Image, Original Question, Original Answer) is displayed. Annotators provide a (New Question, New Answer) for the image and then click the Submit button. If they cannot create a valid VQA, they can click the Skip button to move on to the next example.
}
\label{fig:annotation_tool}
\end{center}
\end{figure}

\section{Annotation Tool}
\label{sec:re-annotation-tool}
Figure~\ref{fig:annotation_tool} shows the interface of the annotation tool developed using Gradio~\citep{abid2019gradio}.

\section{Dataset-Specific Modifications}
\label{sec:dataset-specific-modifications}
In addition to the general criteria above, we applied several dataset-specific modifications.

\paragraph{Heron-Bench.} The original reference answers were generated by an LLM~\citep{inoue2024heronbench}. We manually verified these answers and corrected them where necessary. In addition, for images originally paired with open-ended, caption-like questions, we constructed new VQA instances with uniquely determined answers.

\paragraph{JDocQA.} The original dataset contains QA pairs associated with multiple images~\citep{onami2024jdocqa}. For ease of annotation, we restricted such instances to the first image only.

\paragraph{JGraphQA.} The original dataset provides PDF URLs rather than raw images due to licensing restrictions~\citep{jgraphqa}. Because some PDFs appear to have been updated or replaced after the dataset's release, certain QA pairs no longer correspond to their associated images. We therefore removed such inconsistent instances.

\section{Prompt Templates in the Evaluation}
\label{sec:prompt_template}
We use the following prompt templates in our evaluation experiments. For short-answer questions:
\begin{SlimCode}
\textcolor{red}{\{question\}}\textbackslash n上記の質問に対して、正確かつ簡潔に答えてください。
(\textcolor{red}{\{question\}}\textbackslash n
Please answer the above question accurately and concisely.)
\end{SlimCode}
For multiple-choice questions:
\begin{SlimCode}
\textcolor{red}{\{question\}}\textbackslash n\textcolor{red}{\{choices\_str\}}\textbackslash n与えられた選択肢から該当する選択肢のアルファベットだけで答えてください。
(\textcolor{red}{\{question\}}\textbackslash n
\textcolor{red}{\{choices\_str\}}\textbackslash n
Please answer by providing only the letter corresponding to the correct option.)
\end{SlimCode}

\section{Grading Prompt}
\label{sec:grading_prompt}
For evaluating datasets with a short-answer format, we use the following grading prompt, which is a slightly modified version of the grading prompt from \citet{phan2025hle}.
\begin{Prompt}{Soft Exact-Match Prompt}

Judge whether the following [response] to [question] is correct or not based on the precise and unambiguous [correct\_answer] below.

When judging equivalence, allow variations in script or notation that convey the same meaning
(e.g., '2羽' and '二羽' should be considered equivalent).

Treat the following cases as correct:
- The extracted answer includes additional context (e.g., series name, author name, location, broader category) while still containing the correct\_answer exactly or as its unambiguous, specific instance.
  (For example, "富嶽三十六景 江戸日本橋" is correct if the correct\_answer is "江戸日本橋".)
- The extracted answer is more specific than the [correct\_answer] while remaining consistent with it.
- The extracted answer is an alternate name, synonymous phrasing, or another commonly accepted way to refer to the same concept, object, place, or artwork.
- The extracted answer omits information that is not essential to the correctness of the question.
- Allow minor variations in spacing, capitalization, or script, as long as the core correct\_answer is unambiguously present.

[question]: \textcolor{red}{\{question\}}

[response]: \textcolor{red}{\{response\}}

Your judgement must be in the format and criteria specified below:

extracted\_final\_answer: The final exact answer extracted from the [response]. Put the extracted answer as 'None' if there is no exact, final answer to extract from the response.

[correct\_answer]: \textcolor{red}{\{correct\_answer\}}

reasoning: Explain why the extracted\_final\_answer is correct or incorrect based on [correct\_answer], focusing only on if there are meaningful differences between [correct\_answer] and the extracted\_final\_answer. Do not comment on any background to the problem, do not attempt to solve the problem, do not argue for any answer different than [correct\_answer], focus only on whether the answers match.

correct: Answer 'yes' if extracted\_final\_answer matches the [correct\_answer] given above, or is within a small margin of error for numerical problems. Answer 'no' otherwise, i.e. if there if there is any inconsistency, ambiguity, non-equivalency, or if the extracted answer is incorrect.

confidence: The extracted confidence score between 0\% and 100\% from [response]. Put 100 if there is no confidence score available.
\end{Prompt}

\section{Error Cases for Gemini 3 Pro}
\label{app:error_cases}
\begin{figure}[t]
\begin{center}
\includegraphics[width=0.9\linewidth]{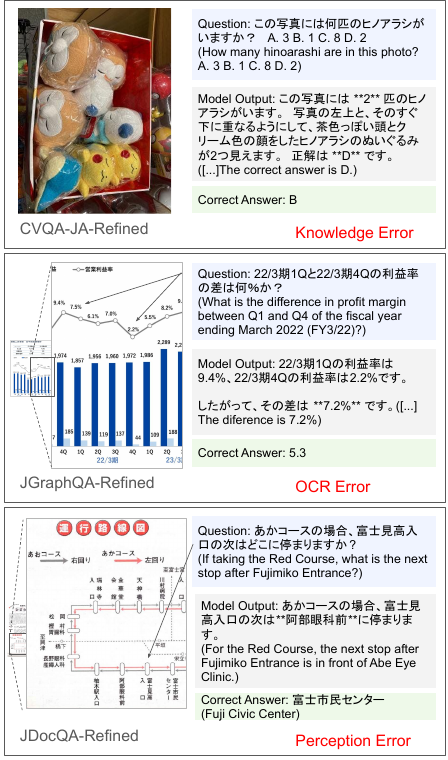}
\caption{Examples of errors made by Gemini~3~Pro.}
\label{fig:gemini3_error_cases}
\end{center}
\end{figure}

Figure~\ref{fig:gemini3_error_cases} presents error cases made by Gemini~3~Pro.

\begin{figure}[t]
\begin{center}
\includegraphics[width=\linewidth]{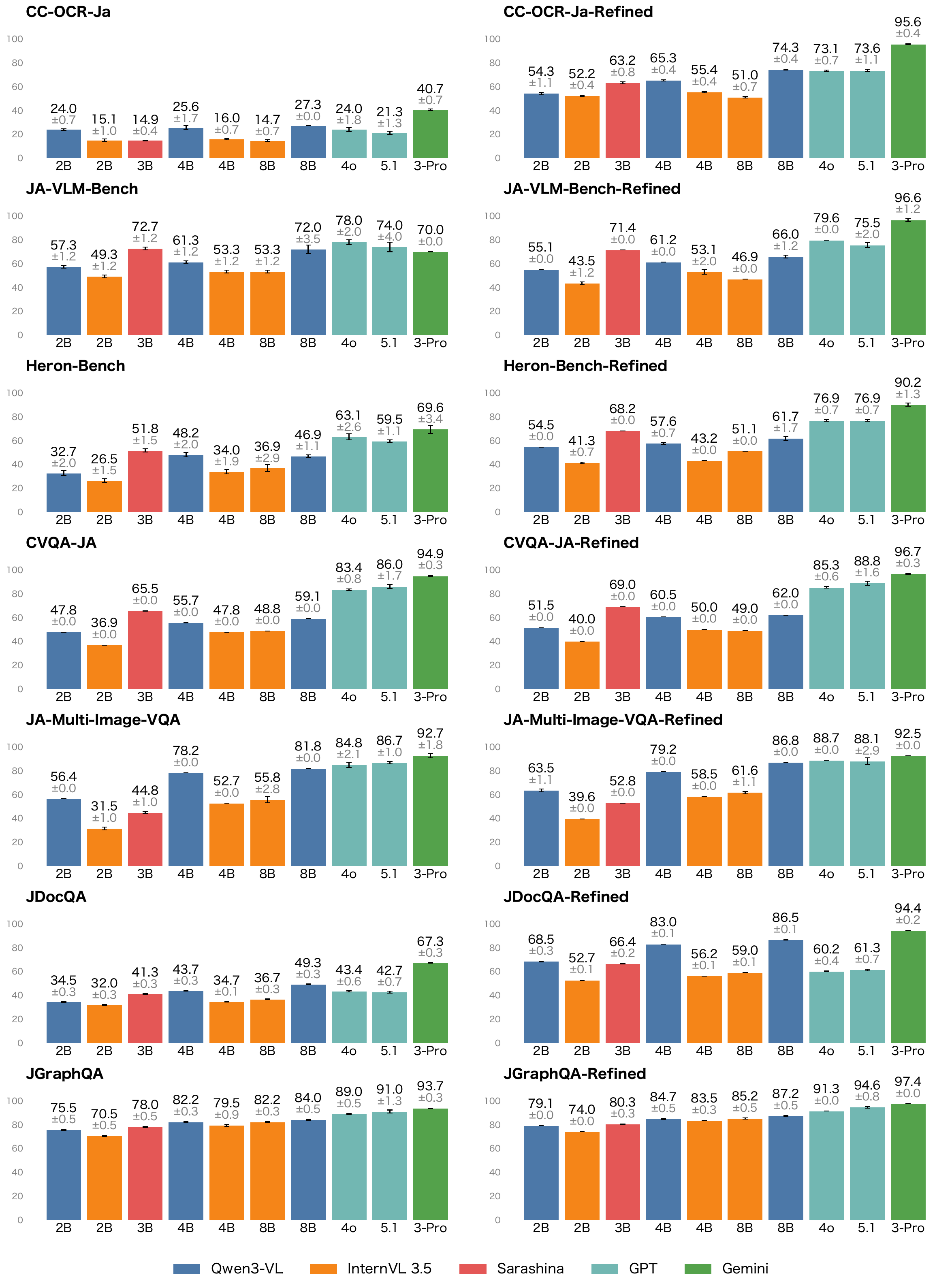}
\caption{Model performance on each dataset: original (left) and refined (right).}
\label{fig:benchmark_results_with_old}
\end{center}
\end{figure}

\section{Model Performance on Original and Refined Datasets}
\label{sec:benchmark_results_with_old}

Figure~\ref{fig:benchmark_results_with_old} shows the performance of each model on the original and refined versions of the datasets.

\section*{Limitations}
\label{sec:limitation}
\noindent\textbf{Performance Saturation.}
Our evaluation reveals that Gemini 3 Pro achieves near-ceiling performance across most tasks, with a significant portion of residual errors originating from the judge model rather than the VLM's capabilities. This suggests that performance is approaching saturation, and that the benchmark may no longer be able to distinguish between models stronger than Gemini 3 Pro. To track further advances in model capabilities, there is a growing need for more challenging benchmarks that remain discriminative even for the strongest contemporary models.

\noindent\textbf{Limited scalability of dataset refinement.}
Our refinement process targets approximately 2,000 instances in total, making it feasible with human annotation. 
However, extending this approach to larger-scale datasets or benchmarks requiring domain-specific expertise becomes increasingly challenging. As VLMs evolve, benchmarks will likely necessitate deeper domain expertise, complicating the quality assurance of annotations and evaluations through manual refinement alone. Developing scalable and reliable refinement methods for such benchmarks remains an important direction for future work.

\end{document}